\documentclass[11pt,a4paper]{article}
\usepackage[hyperref]{acl2020}
\usepackage{float}
\usepackage{times}
\usepackage{soul}
\usepackage{url}
\usepackage[utf8]{inputenc}
\usepackage{graphicx}
\usepackage{amsmath}
\usepackage{amsthm}
\usepackage{amsfonts}
\usepackage{booktabs}
\usepackage{algorithm}
\usepackage{algorithmic}
\usepackage{color}
\usepackage{wrapfig}
\usepackage{flushend}

\renewcommand{\algorithmiccomment}[1]{\bgroup\hfill//~#1\egroup}
\usepackage{microtype}

\aclfinalcopy 
\def\aclpaperid{16} 

\title{Adaptive Dialog Policy Learning with Hindsight and User Modeling}

\author{Yan Cao\textsuperscript{1}\quad Keting Lu \textsuperscript{2}\quad 
    Xiaoping Chen\textsuperscript{1}\quad Shiqi Zhang \textsuperscript{3}\\
   \textsuperscript{1}School of Computer Science, University of Science and Technology of China\\
   \textsuperscript{2}Commercialization Recommending Researching Department, Baidu Inc.\\
   \textsuperscript{3}Department of Computer Science, SUNY Binghamton \\
}

\date{}

\begin{document}
\maketitle
\begin{abstract}
Reinforcement learning methods have been used to compute dialog policies from language-based interaction experiences. 
Efficiency is of particular importance in dialog policy learning, because of the considerable cost of interacting with people, and the very poor user experience from low-quality conversations.
Aiming at improving the efficiency of dialog policy learning, we develop algorithm LHUA (Learning with Hindsight, User modeling, and Adaptation) that, for the first time, enables dialog agents to adaptively learn with hindsight from both simulated and real users. 
Simulation and hindsight provide the dialog agent with more experience and more (positive) reinforcements respectively. 
Experimental results suggest that, in success rate and policy quality, LHUA outperforms competitive baselines from the literature, including its no-simulation, no-adaptation, and no-hindsight counterparts. 
\end{abstract}

\begin{figure*}[htbp] 
\begin{center}
   \includegraphics[width=15cm]{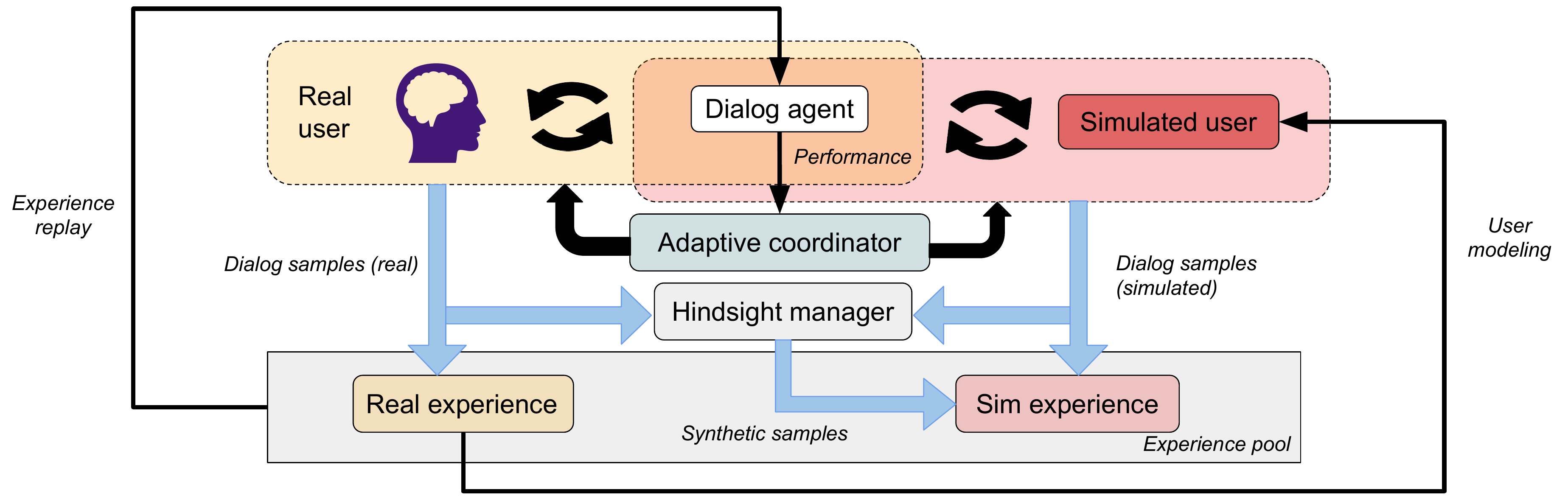} 
\end{center}
\vspace{-.5em}
\caption{An overview of LHUA. 
A \emph{dialog agent} interacts with both real and simulated users while learning a dialog policy from this interaction experience. 
A \emph{simulated user} is modeled using real dialog samples, and interacting with this simulated user provides the dialog agent with simulated dialog samples. 
An \emph{adaptive coordinator} learns from the dialog agent's recent performance to adaptively assign one user (real or simulated) for the dialog agent to interact with. 
A \emph{hindsight manager} manipulates both real and simulated dialog samples (of mixed qualities) to ``synthesize'' successful dialog samples.}
\label{fig:overview}
\vspace{-1em}
\end{figure*}


\section{Introduction}

Dialog systems have enabled intelligent agents to communicate with people using natural language. 
For instance, virtual assistants, such as Siri, Echo, and Cortana, have been increasingly popular in daily life. 
We are particularly interested in goal-oriented dialog systems, where the task is to efficiently and accurately exchange information with people, and the main challenge is on the ubiquitous ambiguity in natural language processing (spoken or text-based). 
Goal-oriented dialog systems typically include components for language understanding, dialog management, and language synthesis, while sometimes the components can be constructed altogether, resulting in end-to-end dialog systems~\cite{bordes2016learning,williams2016end,young2018augmenting}. 
In this paper, we focus on the problem of policy learning for dialog management.

Reinforcement learning (RL) algorithms aim at learning action policies from trial-and-error experiences~\cite{sutton2018reinforcement}, and have been used for learning dialog policies~\cite{young2013pomdp,levin1997learning}. 
Deep RL methods (e.g.~\cite{mnih2013playing}) have been developed for dialog policy learning in dialog domains with large state spaces. 
While it is always desirable for RL agents to learn from the experiences of interacting with the real world, such interactions can be expensive, risky, or both in practice. 
Back to the context of dialog systems, despite all the advances in RL (deep or not), dialog policy learning remains a challenge.
For instance, interacting with people using natural language is very costly, and low-quality dialog policies produce very poor user experience, which is particularly common in early learning phases. 
As a result, it is critical to develop sample-efficient RL methods for learning high-quality dialog policies with limited conversational experiences.

In this paper, we develop an algorithm called LHUA (Learning with Hindsight, User modeling, and Adaptation) for sample-efficient dialog policy learning. 
LHUA, for the first time, enables a dialog agent to simultaneously learn from real, simulated, and hindsight experiences, which identifies the key contribution of this research. 
Simulated experience is generated using learned user models, and hindsight experience (of successful dialog samples) is generated by manipulating dialog segments and goals of the (potentially many) unsuccessful samples. 
Dialog experience from simulation and hindsight respectively provide more dialog samples and more positive feedback for dialog policy learning. 
To further improve the sample efficiency, we develop a meta-agent for LHUA that adaptively learns to switch between real and simulated users in the dialog-based interactions, which identifies the secondary contribution of this research. 
An overview of LHUA is shown in Figure~\ref{fig:overview}.

Experiments were conducted using a realistic Movie-ticket booking platform~\cite{li2017end}. 
LHUA has been compared with state-of-the-art methods~\cite{peng-etal-2018-deep,Lu_2019,Su_2018} in dialog policy learning tasks. 
Results suggest that ablations of LHUA produce comparable (or better) performances in comparison to competitive baselines in success rate, and LHUA as a whole performed the best.

\section{Related Work}
\label{sec:related}

In this section, we summarize three different ways of improving the efficiency of dialog policy learning (namely user modeling, hindsight experience replay, and reward shaping), and qualitatively compare them with our methods.

Researchers have developed ``two-step'' algorithms that first build user models through supervised learning with real conversational data, and then learn dialog policies by interacting with the simulated users~\cite{schatzmann2007agenda,li2016user}. 
In those methods, user modeling must be conducted offline before the start of dialog policy learning.
As a result, the learned policies are potentially biased toward the historical conversational data. 
Toward online methods for dialog policy learning, researchers have developed algorithms for simultaneously constructing models of real users, and learning from the simulated interaction experience with user models~\cite{su-etal-2016-line,lipton2016efficient,zhao-eskenazi-2016-towards,Williams_2017,Dhingra_2017,li2017end,liu2017iterative,peng-etal-2017-composite,wu2019switch}. 
Those methods enable agents to simultaneously build and leverage user models in dialog policy learning. 
However, the problem of learning high-quality user models by itself can be challenging. 
Our algorithms support user modeling, while further enabling agents to adaptively learn from both hindsight and real conversations.

In comparison to many other RL applications, goal-oriented dialog systems have very sparse feedback from the ``real world'' (human users), where one frequently cannot tell dialogs being successful or not until reaching the very end. 
Positive feedback is even rarer, when dialog policies are of poor qualities. 
Hindsight experience replay (HER)~\cite{andrychowicz2017hindsight} methods have been developed to convert unsuccessful trials into successful ones through goal manipulation. 
The ``policy learning with hindsight'' idea has been applied to various domains, including dialog~\cite{Lu_2019}. 
Our methods support the capability of learning from hindsight experience, while further enabling user modeling and learning from simulated users.

Within the dialog policy learning context, reward shaping is another way of providing the dialog agents with extra feedback, where a dense reward function can be manually designed~\cite{su2015reward}, or learned~\cite{su-etal-2016-line}. 
Researchers also developed efficient exploration strategies to speed up the policy learning process of dialog agents, e.g.,~\cite{pietquin2011sample,lagoudakis2003least}. 
Those methods are orthogonal to ours, and can potentially be combined to further improve the dialog learning efficiency. 
In comparison to all methods mentioned in this section, LHUA is the first that enables dialog policy learning from real, simulated, and hindsight experiences simultaneously, and its performance is further enhanced through a meta-policy for switching between interactions with real and simulated users.

\section{Background}
In this section, we briefly introduce the two building blocks of this research, namely Markov decision process (MDP)-based dialog management, and Deep Q-Network (DQN).

\subsection{MDP-based Dialog Management}
Markov Decision Processes (MDPs) can be specified as a tuple \textit{$<\mathcal{S,A},T,\mathcal{R},s_0>$}, where $\mathcal{S}$ is the state set, $\mathcal{A}$ is the action set, $T$ is the transition function, $\mathcal{R}$ is the reward function, and $s_0$ is the initial state. 
In MDP-based dialog managers, dialog control can be modeled using MDPs for selecting language actions.
\textit{$s\in\mathcal{S}$} represents the current dialog state including the agent’s last action, the user’s current action, the distribution of each slot, and other domain variables as needed. 
\textit{$a\in\mathcal{A}$} represents the agent’s response.
The reward function $\mathcal{R}:\mathcal{S}\times\mathcal{A}\to \mathbf{R}$ gives the agent a big bonus in successful dialogs, a big penalty in failures, and a small cost in each turn.

Solving an MDP-based dialog management problem produces $\pi$, a dialog policy. 
A dialog policy maps a dialog state to an action, $\pi: \mathcal{S} \to \mathcal{A}$, toward maximizing the discounted, accumulative reward in dialogs, i.e., $R_t =\sum^{\infty}_{i=t}\gamma^{i-t}r_i$, where $\gamma\in[0, 1]$ is a discount factor that specifies how much the agent favors future rewards.

\subsection{Deep Q-Network}
Deep Q-Network (DQN)~\cite{mnih2015human} is a model-free RL algorithm.
The approximation of the optimal Q-function, $Q^*=Q(s,a;\theta)$, is used by a neural network, where $a$ is an action executed at state $s$, and $\theta$ is a set of parameters. 
Its policy is defined either in a greedy way: $\pi_Q(s) = argmax_{a\in \mathcal{A}} Q(s, a; \theta )$ or being $\epsilon$-greedy, i.e., the agent takes a random action in probability $\epsilon$ and action $\pi_Q(s)$ otherwise.
The loss function for minimization in DQN is usually defined using TD-error:
\begin{equation}
    \mathcal{L}=\mathbf{E}_{s,a,r,s'}[(Q(s,a;\theta)-y)^2],
\end{equation}
where $y=r+\gamma max_{a'\in \mathcal{A}}Q(s',a';\theta)$.

To alleviate the problem of unstable or non-convergence of Q values, two techniques are widely used. 
One is called \textit{target network} whose parameters are updated by $\theta$ once every many iterations in the training phase.
The other technique is \textit{experience replay}, where an experience pool $\varepsilon$ stores samples, each in the form of $(s_t,a_t,r_t,s_{t+1})$.
It randomly selects small batches of samples from $\varepsilon$ each time during training.
Experience replay can reduce the correlation between samples, and increases the data efficiency.

\section{Algorithms}
\label{sec:alg}

In this section, we first introduce Learning with Hindsight, and User modeling (\textbf{LHU}), and then present LHU with Adaptation (\textbf{LHUA}), where algorithms LHU and LHUA point to the main contribution of this research. 

LHU, for the first time, enables a dialog agent to learn dialog policies from \textbf{three dialog sources}, namely real users, simulated users, and hindsight dialog experience. 
More specifically, a real user refers to the human who converses with the dialog agent, and a simulated user refers to a learned user model that captures real users' interactive behaviors with our dialog agent. 
In this way, a simulated user is used for generating ``human-like'' dialog experience for speeding up the process of dialog policy learning. 
The last dialog source of ``hindsight dialog experience'' is used for creating many \emph{successful} dialog samples using both successful and unsuccessful dialog samples, where the source samples are from both real and simulated users. 
Different from ``simulated users'' that generate dialog samples of mixed qualities, hindsight experience produces only successful (though not real) dialog samples, which is particularly useful for dialog policy learning at the early phase due to the very few successful samples. 

Among the three dialog sources, hindsight experience is ``always on'', and synthesizes dialog samples throughout the learning process. 
The ``real'' and ``simulated'' dialog sources bring in the selection problem: \emph{At a particular time, from which source should the agent obtain dialog experience for policy learning? }
The ``adaptation'' capability of LHUA aims at enabling the dialog agent to learn to, before starting a dialog, select which user (real or simulated) to interact with.

\subsection{Learning with Hindsight, and User Modeling}
\label{sec:lhu}

In this subsection, we focus on two components of LHUA, including user modeling, and hindsight management, which together form LHU, an ablation algorithm of LHUA. 
The two components' shared goal is to generate additional dialog experience (simulated and hindsight experiences respectively) to speed up dialog policy learning.

\paragraph{Dialog (Sub)Goal and Segmentation}
Goal-oriented dialog agents help users accomplish their goals via language-based multi-turn communications. 
Goal $G$ includes a set of constraints $C$ and a set of requests $R$, where $G=(C,R)$. 
Consider a service request ``\emph{I'd like to purchase one ticket of Titanic for this evening. Which theater is available?}''  
In this example, the goal is of the form:
\begin{align*}
  G=\big( &C=[ticket=one, time=eve,\\
& \qquad movie=titanic], \\
  &R=[theater=?]\big)
\end{align*}

We define $G'$ as a subgoal of $G=(C,R)$: $G'=(C',R')$, where $C'\subseteq C$, $R'\subseteq R$, and  $G'$ cannot be empty. 
Continuing the ``titanic'' example, one of its subgoals is 
\begin{align*}
G'=\big( &C'=[ticket=one,movie=titanic],\\
&R'=\emptyset\big).
\end{align*}

Given an intact dialog $D$, we say $D_{seg}$ is a segment of $D$, if $D_{seg}$ includes a consecutive sequence of turns of $D$.
With the concepts of dialog segment and subgoal, we introduce two segment sets (head and tail), which are later used in \textit{hindsight manager}.
A head segment set $\Omega$ consists of dialog segments $D_{head}$ that include the early turns in the intact dialog with the corresponding completed subgoal $G'$.
\begin{equation}
    \Omega = \{(D_{head},G') \}
\end{equation}

We use function $HeadSegGen$ to collect a head segment set $\Omega$ during dialog interactions.
$HeadSegGen$ receives a dialog segment $D_{seg}$, and a goal $G$, then checks all subgoals of $G$, and finally outputs pairs $(D_{seg},G')$ where $D_{seg}$ accomplishes subgoal $G'$ of $G$.

A tail segment set $\Gamma$ consists of dialog segments $D_{tail}$ that include the late turns in the intact dialog with the corresponding completed subgoal $G'$.
\begin{equation}
    \Gamma =\{(D_{tail},G')\}
\end{equation}
Function $TailSegGen$ is implemented to generate tail segments after interactions terminate.
It receives a dialog $D$, a goal $G$ and a corresponding head segment $\Omega$.
If the dialog $D$ accomplishes the goal $G$, for each pair $(D_{head},G')$ from the head segment set $\Omega$, $TailSegGen$ outputs a corresponding pair $(D\ominus D_{head},G')$, where $D_1\ominus D_2$ produces a dialog segment by removing $D_2$ from $D_1$.

\paragraph{Hindsight Manager}
Given head and tail segment sets ($\Omega$ and $\Gamma$), \textit{hindsight manager} is used for stitching two tuples, $(D_{head},G'_{head})$ and $(D_{tail},G'_{tail})$, respectively to ``synthesize'' successful dialog samples.
There are two conditions for synthesization: 
\begin{enumerate}
    \item The two subgoals from head and tail segments are identical, $ G'_{head} == G'_{tail} \label{Syn-con1}$, and
    \item The last state of $D_{head}$, $s_{last}$, and the first state of $D_{tail}$, $s'_{first}$, are of sufficient similarity. 
\end{enumerate}

We use \textit{KL Divergence} to measure the similarity between two states:
\begin{equation}
\begin{aligned}
        D_{KL}(s_{last} || s'_{first}) \leq \delta \label{condition2}
\end{aligned}
\end{equation}
where $\delta\in R$ is a threshold parameter.
We implement a function to synthesize successful dialog samples as hindsight experience for dialog policy learning, as follows:
\begin{equation}
    D_{hind} \leftarrow HindMan(\delta,\Omega,\Gamma)
\end{equation}

$HindMan$ takes a threshold $\delta$, a head segment set $\Omega$, and a tail segment set $\Gamma$. It generates successful dialog samples $D_{hind} $ that satisfy the above two conditions of synthesization.

\begin{algorithm}[t] \footnotesize
\caption{Algorithm LHU}
\label{algorithm2.1:LHU} 
\textbf{Input}: 
$K$, the times of interactions with the simulated user; 
$\delta$, KL-divergence threshold
\vspace{.5em}
\\ 
\textbf{Output}: 
the success rate $SR^{Dlg}$, and average rewards $R^{Dlg}$ of $agent^{Dlg}$;
$Q(\cdot)$ for $agent^{Dlg}$ 
\begin{algorithmic}[1] 
\STATE Initialize $Q(s,a;\theta_Q)$ of $agent^{Dlg}$ and $M(s,a;\theta_M)$ of the simulated user via pre-training on human conversational data
\STATE Initialize experience replay buffers $B^R$ and $B^S$ for the interaction of $agent^{Dlg}$ with real and simulated users
\STATE Initialize head and tail dialog segment sets:\\ \begin{center} $\Omega\leftarrow\emptyset$, and $\Gamma\leftarrow\emptyset$ 
\end{center}
\STATE Collect initial state, $s$, by interacting with a real user following goal $G^{Real}$ \label{user:start}
        \STATE Initialize $D^{Real}\leftarrow \emptyset $ for storing dialog turns (real)
        \WHILE[\textcolor{blue}{Start a dialog with real user}]{$s\notin$ term} \label{real loop}
            \STATE Select $a\leftarrow argmax_{a'}Q(s,a';\theta_Q)$, and execute $a$
            \STATE Collect next state $s'$, and reward $r$
            \STATE Add dialog turn $d=(s,a,r,s')$ to $B^R$ and $D^{Real}$
            \STATE $\Omega \leftarrow \Omega \cup HeadSegGen(D^{Real},G^{Real})$ \label{user: headseggen}
            \STATE $s\leftarrow s'$
        \ENDWHILE
        \STATE $ \Gamma \gets \Gamma \cup TailSegGen(D^{Real},G^{Real},\Omega)$ \label{user: tailseggen}\label{user:end}
        \FOR[\textcolor{blue}{$K$ interactions with simulated user}]{$k=1:K$}\label{sim interactions}
            \STATE Sample goal $G^{Sim}$, and initial state $s$
            \STATE Initialize $D^{Sim}\leftarrow \emptyset$ for storing dialog turns (sim)
            \WHILE[\textcolor{blue}{The $k^{th}$ dialog with sim user}]{$s\notin$ term}\label{sim loop}
                \STATE $a \! \leftarrow \! argmax_{a'}Q(s,a';\theta_Q)$, and execute $a$
                \STATE Collect next state $s'$, and reward $r$ from $M(s,a;\theta_M)$
                \STATE Add dialog turn $d=(s,a,r,s')$ to $B^S$ and $D^{Sim}$ \label{simsample:model}
                \STATE $\Omega \leftarrow \Omega \cup HeadSegGen(D^{Sim},G^{Sim})$ \label{sim: headseggen}
                \STATE $s\leftarrow s'$
            \ENDWHILE
            \STATE $ \Gamma \gets \Gamma \cup TailSegGen(D^{Sim},G^{Sim},\Omega)$ \label{sim:tailseggen}
        \ENDFOR\label{sim:end}
        \STATE Synthesize hindsight experience, and store it in $B^S$:
        $D_{hind} \! \leftarrow \!\! HindMan(\delta,\Gamma,\Omega)$
        \label{hindsight: HindMan}
        \COMMENT{\textcolor{blue}{Hindsight Manipulation}}
        \STATE Calculate the success rate $SR^{Dlg}$ and average rewards $R^{Dlg}$ of total interactions \label{training result}
        \STATE Randomly sample a minibatch from both $B^R$ and $B^S$, and update $agent^{Dlg}$ via DQN \label{policy learning}
        \COMMENT{\textcolor{blue}{$agent^{Dlg}$ training}} 
        \STATE Randomly sample a minibatch from $B^R$, and update simulated user via SGD \label{user modeling} 
        \COMMENT{\textcolor{blue}{User modeling}}
        \vspace{-1em}
        \RETURN $SR^{Dlg}$, $R^{Dlg}$, $Q(\cdot)$
\end{algorithmic}
\end{algorithm}

\paragraph{Dialog with Simulated Users}
In dialog policy learning, dialog agents can learn from interactions with real users, where the generated real experience is stored in reply buffer $B^R$. 
To provide more experience, we develop a simulated user for generating simulated dialog experience to further speed up the learning of dialog policies.

The simulated user is of the form: 
$$
  s',r \leftarrow M(s,a;\theta_M)
$$
where, $M(s,a;\theta_M)$ takes the current dialog state $s$ and the last dialog agent action $a$ as input, and generates the next dialog state $s'$, and reward $r$. 
$M$ is implemented by a Multi-Layer Perceptron (MLP) parameterized by $\theta_M$, and refined via stochastic gradient descent (SGD) using real experience in $B^R$ to improve the quality of simulated experience.

Simulated experience generated from interactions between the dialog agent and the simulated user is stored in simulated replay buffer $B^S$, which is also manipulated by \textit{hindsight manager} to synthesize hindsight experience.

\paragraph{The LHU Algorithm}

Algorithm~\ref{algorithm2.1:LHU} presents the learning process, where our dialog agent interacts with a real user for one dialog, and a simulated user for $k$ dialogs. 
In addition to parameter $k$, there is a \emph{KL-divergence} threshold $\delta$ as a part of the input.
We refer to this algorithm using LHU($k$).

Algorithm~\ref{algorithm2.1:LHU} starts with an initialization of the dialog agent's real and simulated experience replay buffers ($B^R$ and $B^S$ respectively), the model of the simulated user, $M(\theta_M)$, and two segment sets for \textit{hindsight manager} ($\Omega$ and $\Gamma$ respectively).
In the first \textit{while} loop (starting in Line~\ref{real loop}), the dialog agent interacts with a real user and stores the real experience in $B^R$.
Then, $k$ dialogs with the simulated user are conducted in the \textit{for} loop, where simulated experience is stored in $B^S$.
During interactions with both real and simulated users, head and tail segment sets are simultaneously collected (Lines 21 and 24).
After all dialog interactions end, the \textit{hindsight manager} is used to synthesize successful dialog samples and store them in $B^S$.
Finally, the dialog agent is trained on $B^R$ and $B^S$, and the simulated user is trained on $B^R$.

The output of Algorithm~\ref{algorithm2.1:LHU} is used in the next section, where we introduce how to further enable the dialog agent to learn a meta-policy for adaptively determining to which user (real or simulated) to interact with.

\subsection{LHU with Adaptation (LHUA)}
\label{sec:lhua}

Adaptively determining which user (real or simulated) the LHU agent should interact with can further speed up the dialog policy learning process. 
The idea behind it is that, if a simulated user can generate high-quality, realistic dialog experience, interactions with the simulated user should be encouraged. 
To enable this adapative ``switching'' behaviors, we develop an \textit{adaptive coordinator} that learns a meta-policy for selecting between real and simulated users for collecting interaction experience. 
We learn this adaptive coordinator using reinforcement learning, producing the LHUA algorithm, which is described next. 


\begin{algorithm}[t] \footnotesize
\caption{LHU with Adaptation (LHUA)}
\label{algorithm2.2:LHUA} 
\textbf{Input}:  
$H$, the max length of adaptation episode; 
$\delta$, $KL$-divergence threshold; 
$N$, training times\vspace{.5em}
\\ 
\textbf{Output}: 
$\Pi$, the dialog policy;
\begin{algorithmic}[1] 
\STATE Initialize $A(s^A,k;\theta_A)$ of $agent^{Adp}$, and replay buffer $B^A$ as empty
\FOR{$i=1:N$}
    \STATE Initialize adaptation state $s^{A}$ using Eqn.~\ref{equ1} \label{adp:state initialize} 
    \STATE Initialize turn counter $h$: $h=0$ \label{adp: turn counter}
    \WHILE{$h\leq H$}\label{adp: H}
        \STATE Select action $k$:\label{adp:select}
        $k\leftarrow argmax_{k'}A(s^A,k';\theta_A)$ \label{adp:action}
        \STATE Execute action $k$:
        \vspace{-0.5em}
        $$SR^{Dlg}, R^{Dlg}, Q(\cdot) \leftarrow LHU^{\ref{algorithm2.1:LHU}}(k,\delta)$$
        \vspace{-1.5em}
        \STATE Collect reward $r^A$ via Eqn.~\ref{adp reward fun}, and next adaptation state $\hat{s}^A$ using Eqn.~\ref{equ1}  \label{adp:state and reward}
        \STATE $B^A\leftarrow B^A \cup (s^{A}, k, r^{A},\hat{s}^{A})$, $s^{A}\leftarrow \hat{s}^{A}$, and $h\leftarrow h+1$ \label{adp: state update}
    \ENDWHILE
    \STATE Sample a minibatch from $B^A$, and update $\theta_A$ via DQN \label{adp: update}
    \vspace{-1em}
\ENDFOR
\STATE for all $s\in \mathcal{S}$:
$ \Pi (s)\leftarrow argmax_{a'}Q(s,a';\theta_Q)$
\RETURN $\Pi(\cdot)$
\end{algorithmic}
\end{algorithm}

\paragraph{State}
In each turn of interaction with the LHU agent, \textit{adaptive coordinator} updates the adaptation state $s^A$ using the equation below:
\begin{equation}\footnotesize
s_i^A=
\begin{cases}
      [0,0,0,0] &  {i=0} \\
      [SR_i,R_i,SR_i-SR_{i-1},R_i-R_{i-1}]& {i > 0 } \label{equ1}
\end{cases}
\end{equation}
where $SR_i$ and $R_i$ are respectively average success rate and rewards from LHU agent's training performance at $i^{th}$ episode.
In practice, $R$ is normalized to have values between $0$ and $1$, same as $SR$.
This form of adaptation state provides accessible information on different training phrases to represent LHU agent 's current performance.

\paragraph{Action}
Based on the state $s^A$, \textit{adaptive coordinator} chooses action $k$ to determine, after each dialog with the real user, how many dialogs should be conducted with the simulated user. 
The value of action $k$ ranges from 1 to $K$.

\paragraph{Reward}
\textit{Adaptive coordinator} receives immediate rewards after executing an action k (i.e. LHU(k)) each time.
We use success rate increment of LHU agent to design the reward function, as shown below:
\begin{equation}
    r^A_i=\frac{SR_i-SR_{i-1}}{SR_{i}} \cdot \frac{k_i}{L_i}\quad (0<i\leq H) \label{adp reward fun}
\end{equation}
where $k_i$ is the $i^{th}$ action chosen by \textit{adaptive coordinator}, and $L_i$ means the total number of times of interactions with both real and simulated users.
From the above we can know $L_i=k_i+1$.
Reward is continuously harvested, until the $H^{th}$ turn.

Due to the continuous state space, the approximated value function of \textit{adaptive coordinator} is implemented using a two-layer fully connected nerual network, $A(s^A,k;\theta_A)$, parameterized by $\theta_A$.
Interactions between the \textit{adaptive coordinator} and the LHU agent start with an initial state.
In each turn, the \textit{adaptive coordinator} obtains the state $s^A$ using Eqn.~\ref{equ1}, and selects the action $k$ via $\epsilon$-greedy policy to execute.
Then, the current training performance of LHU agent is used for acquiring the reward $r^A$ using Eqn.~\ref{adp reward fun}, and updating the next state $\hat{s}^A$.
Finally, the experience $(s^A,k,r^A,\hat{s}^A)$ is stored for meta-policy learning.
We improve the value function by adjusting $\theta_A$ to minimize the mean-squared loss function.

\paragraph{The LHUA Algorithm}
Algorithm~\ref{algorithm2.2:LHUA} presents the dialog policy learning process, where our dialog agent adaptively learns from both simulated and real users. 
In addition to parameter $\delta$ for KL-divergence threshold, there is parameter $H$ representing the length of one episode for {adaptive coordinator} as a part of the input.

Algorithm~\ref{algorithm2.2:LHUA} starts with an initialization of replay buffer $B^A$ for {adaptive coordinator}, and the value function $A(s^A,k;\theta_A)$.
Before the start of each episode, a turn counter $h$ is initialized as zero for turn counting.
{Adaptive coordinator} interacts with LHU agent for $H$ turns while collecting and saving experience in $B^A$.
At the end of each adaptation episode, we use DQN to update $\theta_A$.

LHUA enables the dialog agent to simultaneously learn from the dialogs with both real and simulated users. 
At the same time, \textit{hindsight manager} manipulates both real and simulated dialog samples to synthesize more successful dialog samples.
Dialog experience from simulation and hindsight respectively provide more dialog samples and more positive feedback for dialog policy learning. 
The \textit{adaptive coordinator} is learned at runtime for adaptively switching between real and simulated users in the dialog policy learning process to further improve the sample efficiency.
So far, LHUA enables dialog agents to adaptively learn with hindsight from both simulated and real users.

\section{Experiment}
\label{sec:exp}

Experiments have been conducted in a dialog simulation platform, called TC-bot~\cite{li2016user,li2017end}.\footnote{To avoid possible confusions, we use ``real user'' to refer to the user directly provided by TC-bot, and use ``simulated user'' to refer to the user model learned by our dialog agents. }
TC-bot provides a realistic simulation platform for goal-oriented dialog system research. 
We use its \emph{movie-ticket booking} domain that consists of 29 slots of two types, where one type is on \textit{search constraints} (e.g., number of people, and date), and the other is on \textit{system-informable} properties that are needed for database queries (e.g., critic rating, and start time). 
The dialog agent has 11 dialog actions, representing the system intent (e.g., confirm question, confirm answer, and thanks).

A dialog is considered successful only if movie tickets are booked successfully, and the provided information satisfies all the user's constraints.
By the end of a dialog, the agent receives a  bonus (positive reward) of $2*L$ if successful, or a penalty (negative reward) of $-L$ for failure, where $L$ is the maximum number of turns allowed in each dialog. 
We set $L=40$ in our experiments. 
The agent receives a unit cost in each dialog turn to encourage shorter conversations.

\paragraph{Implementation Details} 
In line with existing research \cite{peng-etal-2018-deep}, all dialog agents are implemented using Deep Q-Network (DQN). The DQN includes one hidden layer with 80 hidden nodes and ReLU activation, and its output layer of 11 units corresponding to 11 dialog actions. 
We set the discount factor $\gamma = 0.95$.
The techniques of target network and experience replay are applied.
Both $B^R$ and $B^S$ share the buffer size of 5000, and we use uniform sampling in experience replay. 
The target value function is updated at the end of each epoch. 
In each epoch, $Q(\cdot)$ and $M(\cdot)$ are refined using one-step 16-tuple-minibatch update.
We then pre-filled the experience replay buffer with 100 dialogs before training. 
The simulated experience buffer $B^S$ is initialized as empty. 
Neural network parameters are randomly initialized, and optimized using RMSProp~\cite{hinton2012neural}.

The simulated user model, $M(\cdot)$, is a multi-task neural network~\cite{liu2015representation}, and contains two shared hidden layers and three task-specific hidden layers, where each layer has 80 nodes. 
Stitching threshold of \textit{hindsight manager} $\delta$ is set 0.2.
The policy network of \textit{adaptive coordinator} is a single-layer neural network of size 64. 
Parameters $k$ and $H$ are described in Algorithm~\ref{algorithm2.2:LHUA}, and have the value of $k=20$ and $H=8$.

\paragraph{LHUA and Three Baselines}
Our key hypothesis is that adaptively learning from real, simulated, and hindsight experiences at the same time performs better than baselines from the literature. 
To evaluate this hypothesis, we have selected three competitive baselines for goal-oriented dialog policy learning, including DDQ~\cite{Su_2018}, D3Q~\cite{wu2019switch}, and S-HER~\cite{Lu_2019}. 
In implementing the DDQ agent, the ratio of interaction experiences between simulated and real users is ten, which is consistent to the original implementation~\cite{Su_2018}. 
The differences between LHUA and the baseline methods are qualitatively discussed in Section~\ref{sec:related}.

It is necessary to explain how the curves are generated in the figures to be reported. 
For each of the four methods (LHUA and three baselines), we have conducted five ``runs'', where each run include 250 episodes. 
In each run, after every single episode for learning, we let the dialog agent interact with the real user for 50 dialogs. 
We then compute the success rate over the 50 dialogs. 
Each data point in the figure is an average over the five success rates collected from the five runs of each method.

\begin{figure}[t]
\vspace{-1em}
\centering 
\includegraphics[width=0.5\textwidth]{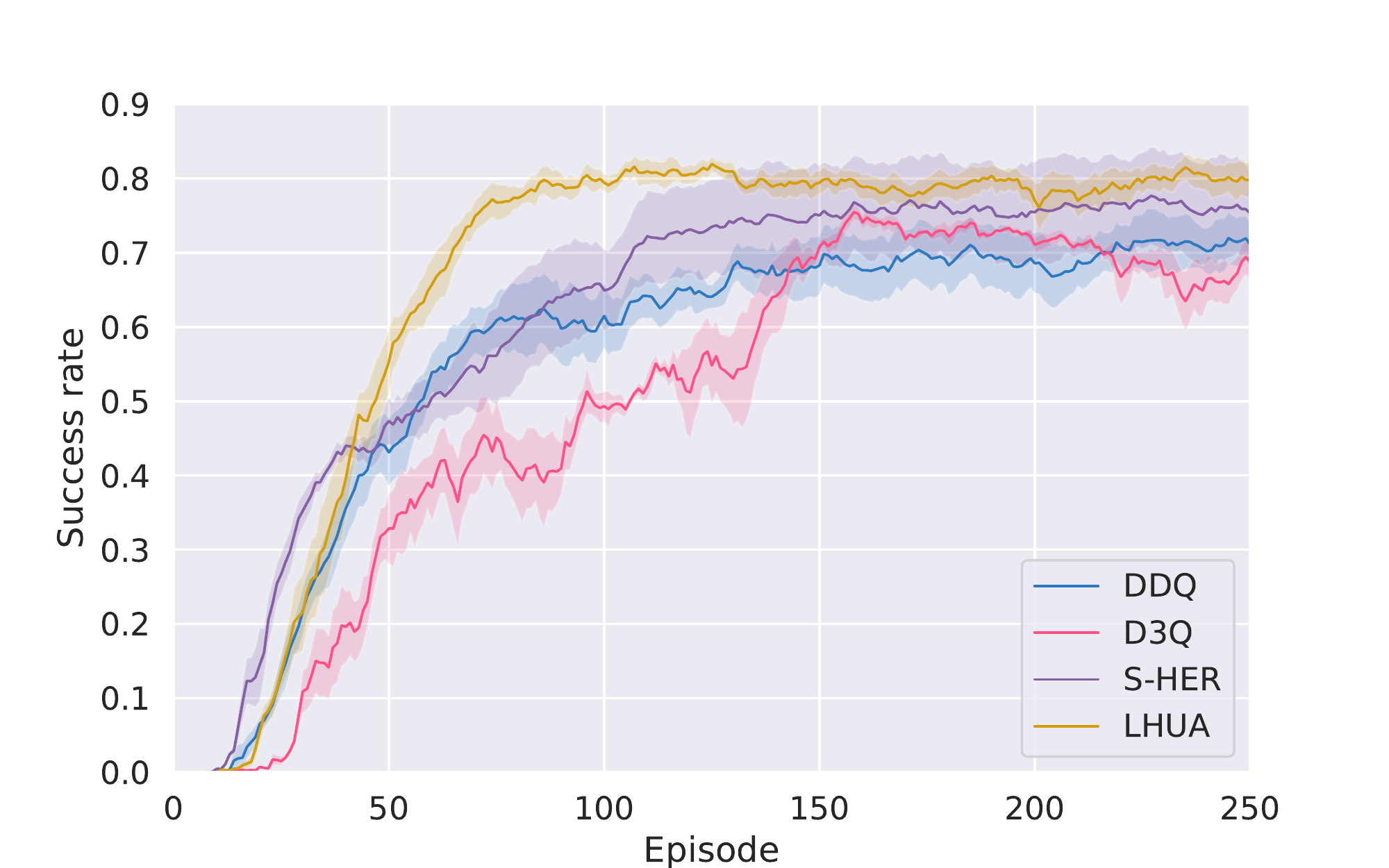}
\vspace{-1.2em}
\caption{The performances of LHUA (ours), and three baseline methods, including DDQ~\cite{Su_2018}, D3Q~\cite{wu2019switch}, and S-HER~\cite{Lu_2019}. We see that, except for the very early phase (first 50 episodes), LHUA outperformed all baselines.} 
\label{fig:mainexp}
\vspace{-.5em}
\end{figure}

Figure~\ref{fig:mainexp} presents the key results of this research on the quantitative comparisons between LHUA and the three baselines. 
We can see that, except for the very early learning phase, LHUA performed consistently better than the three baseline methods. 
In particular, LHUA reached the success rate of 0.75 after about 70 episodes, whereas none of the baselines were able to achieve comparable performance within 150 episodes. 
The gap between LHUA and S-HER in early phase is due to the fact that LHUA needs to learn a user model, which requires extra interaction in early phase. 
Once the user model is of reasonable quality, LHUA is able to learn from the interaction experience with simulated users, and soon (after 45 episodes) LHUA outperformed S-HER.

\begin{figure}[t] 
\vspace{-1em}
\centering \hspace*{-1em}
\includegraphics[width=0.54\textwidth]{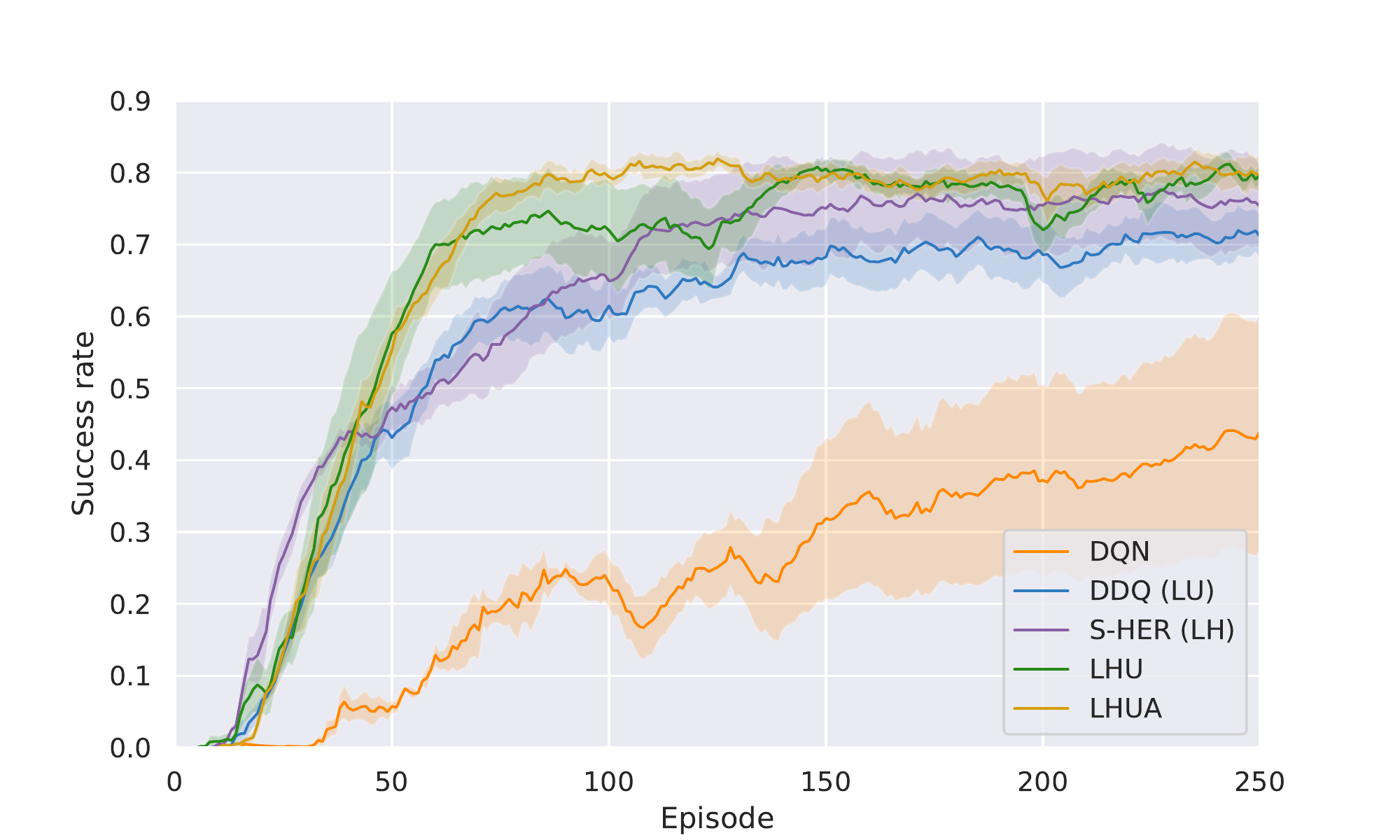}
\caption{Comparisons between LHUA and its ablations: DQN (no hindsight manager, no user modeling, and no adaptive coordinator), DDQ (no hindsight manager, and no adaptive coordinator), S-HER (no user modeling, and no adaptive coordinator), and LHU (no adaptive coordinator). A complete LHUA includes all the components, including DQN (for naive dialog policy learning), hindsight manager, user modeling, and adaptive coordinator. }
\label{fig:exp-ablation}
\end{figure}

\begin{figure}[t] 
\vspace{-1em}
\centering \hspace*{-1.8em}
\includegraphics[width=0.56\textwidth]{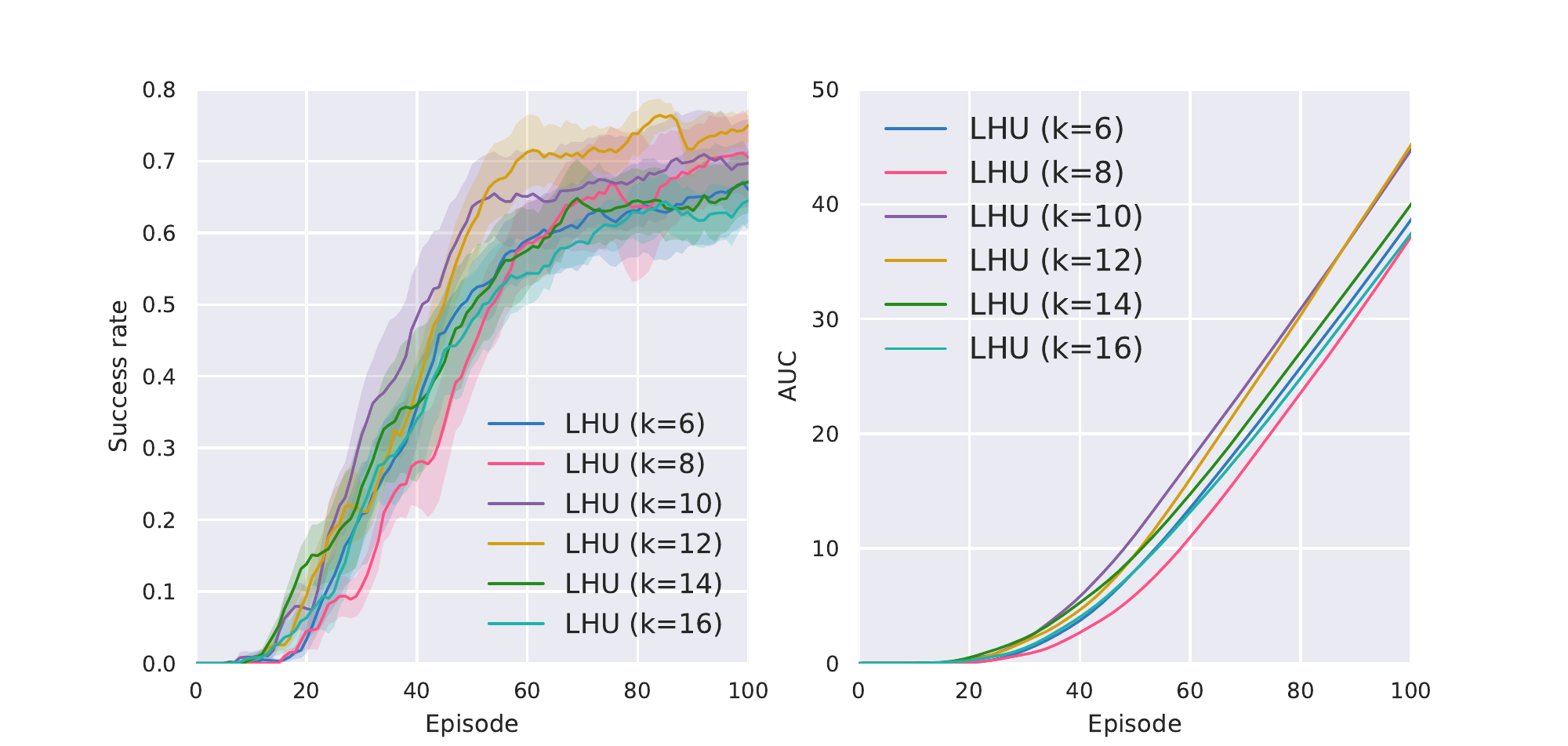}
\caption{Success rate on the left, and Area under Curve (AUC) on the right, where we implemented six different versions of LHU with different $k$ values, ranging from 6 to 16 at an interval of 2. }
\label{fig:lhu}
\end{figure}

\paragraph{LHUA and Its Ablations}
Results reported in Figure~\ref{fig:mainexp} have shown the advantage of LHUA over the three baseline methods. 
However, it is still unclear how much each component of LHUA contributes to its performance. 
We removed components from LHUA, and generated four different ablations of LHUA, including DQN, DDQ (LU, or Learning with User modeling), S-HER (LH, or Learning with Hindsight), LHU, and LHUA.

Figure~\ref{fig:exp-ablation} shows the ablation experiment's results. 
From the results, we see that LHUA performed much better than no-hindsight (LU), and no-user-modeling  (S-HER, or LH) ablations. 
When both ``hindsight'' and ``user modeling'' are activated, there is LHUA's ablation of LHU, which performed better than all the other ablations. 
LHU still cannot generate comparable performance, c.f., LHUA, which justified the necessity of the adapative coordinator. 
It should be noted that performances of two of the ablations have been reported in Figure~\ref{fig:mainexp}. 
We intentionally include their results in Figure~\ref{fig:exp-ablation} for the completeness of comparisons.

\paragraph{Adaptive Coordinator Learning}
Results reported in Figure~\ref{fig:exp-ablation} have shown the necessity of our adaptive coordinator in LHUA. 
In this experiment, we look into the learning process of the adaptive coordinator. 
More specifically, we are interested in how the value of $k$ is selected (see Algorithm~\ref{algorithm2.2:LHUA}). 
We have implemented LHU with six different values of $k$, and their performances are reported in Figure~\ref{fig:lhu}, where the left subfigure is on success rate, and the right is on Area under Curve (AUC). 
The AUC metric has been used for the evaluation of learning speed~\cite{taylor2009transfer,stadie2015incentivizing}.
We see that, in early learning phase 
(within 100 episodes), the $k$ value of 10 produced the best performance overall, though the performance is comparable to that with $k=12$ to some level.

\begin{wrapfigure}{r}{0.27\textwidth}
\vspace*{-8pt}
\includegraphics[width=0.26\textwidth]{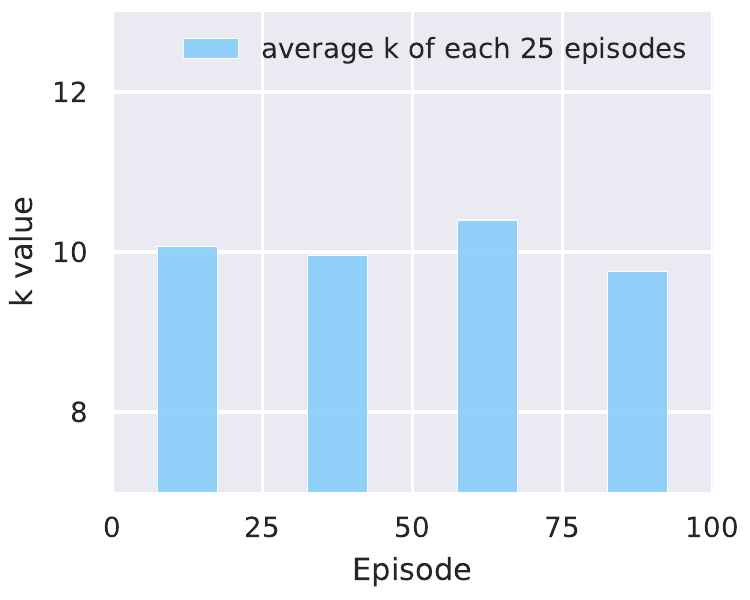}
\vspace{-.5em}
\caption{The $k$ values selected by the \textit{adaptive coordinator} of our LHUA agent}
\vspace{-.5em}
\label{fig:k}
\end{wrapfigure}

Figure~\ref{fig:k} reports the selection of $k$ values by our adaptive coordinator. 
Each bar corresponds to an average over the $k$ values of 25 episodes. 
We see that the value of $k$ was suggested to be around 10 within the first 100 episodes, which is consistent to our observation from the results of Figure~\ref{fig:lhu}. 
The consistency further justified our adaptive coordinator's capability of learning the interaction strategy in switching between real and simulated users. 

\section{Conclusions and Future Work}
In this work, we develop an algorithm called LHUA (Learning with Hindsight, User modeling, and Adaptation) for sample-efficient dialog policy learning. 
LHUA enables dialog agents to adaptively learn with hindsight from both simulated and real users. 
Simulation and hindsight provide the dialog agent with more experience and more (positive) reinforcements respectively. 
Experimental results suggest that LHUA outperforms competitive baselines (including success rate and learning speed) from the literature, including its no-simulation, no-adaptation, and no-hindsight counterparts. 
This is the first work that enables a dialog agent to adaptively learn from real, simulated, and hindsight experiences all at the same time. 

In the future, we plan to evaluate our algorithm using other dialog simulation platform, e.g., PyDial~\cite{ultes2017pydial}, and other testing environments. 
Another direction is to combine other efficient exploration strategies to further improve the dialog learning efficiency.
Finally, we will further consider the noise from language understanding and generation.

\bibliography{acl2020}
\bibliographystyle{acl_natbib}

\appendix
\end{document}